\documentclass[final]{cvpr}

\usepackage{times}
\usepackage{epsfig}
\usepackage{graphicx}
\usepackage{amsmath}
\usepackage{amssymb}

\usepackage{multicol}
\usepackage{booktabs}
\usepackage{multirow}
\usepackage{subfig}

\usepackage[pagebackref=true,breaklinks=true,colorlinks,bookmarks=false]{hyperref}

\begin{document}

\title{Price Suggestion for Online Second-hand Items with Texts and Images}

\author{Liang Han
\thanks{Liang Han is a PhD student at Department of Computer Science, Stony Brook University. He participated in this project as a research intern.}\\
Stony Brook University\\
{\tt\small liahan@cs.stonybrook.edu}
\and
Zhaozheng Yin
\thanks{Corresponding author. Zhaozheng Yin is an associate professor at Department of Computer Science and Department of Biomedical Informatics, Stony Brook University. He participated in this project during his sabbatical.}\\
Stony Brook University\\
{\tt\small zyin@cs.stonybrook.edu}

\and
Zhurong Xia\\
Alibaba Group\\
{\tt\small zhurong.xzr@alibaba-inc.com}

\and
Mingqian Tang\\
Alibaba Group\\
{\tt\small mingqian.tmq@alibaba-inc.com}

\and
Rong Jin\\
Alibaba Group\\
{\tt\small jinrong.jr@alibaba-inc.com}
}

\maketitle


\begin{abstract}
This paper presents an intelligent price suggestion system for online second-hand listings based on their uploaded images and text descriptions. The goal of price prediction is to help sellers set effective and reasonable prices for their second-hand items with the images and text descriptions uploaded to the online platforms. Specifically, we design a multi-modal price suggestion system which takes as input the extracted visual and textual features along with some statistical item features collected from the second-hand item shopping platform to determine whether the image and text of an uploaded second-hand item are qualified for reasonable price suggestion with a binary classification model, and provide price suggestions for second-hand items with qualified images and text descriptions with a regression model. To satisfy different demands, two different constraints are added into the joint training of the classification model and the regression model. Moreover, a customized loss function is designed for optimizing the regression model to provide price suggestions for second-hand items, which can not only maximize the gain of the sellers but also facilitate the online transaction. We also derive a set of metrics to better evaluate the proposed price suggestion system. Extensive experiments on a large real-world dataset demonstrate the effectiveness of the proposed multi-modal price suggestion system.
\end{abstract}

\section{Introduction}
\label{sec:intro}

In recent years, encouraged by the sharing economy, a growing number of people are willing to cash in their second-hand idle assets and consider used products as a cheaper alternative, and second-hand business is heating up. Benefiting from the unprecedentedly booming of the E-commerce, which is fuelled by the rapid prevalence of laptops, smartphones and internet, online second-hand marketplace is becoming an appealing option for people to sell their idle goods, and a number of online second-hand platforms, such as eBay, Xianyu, Mercari and Letgo, have emerged to facilitate the online second-hand trading.

Generally, these online second-hand platforms make no restriction on how the sellers set prices for their listings. However, different from the brand new products, most of the second-hand items listed on the platforms are unique and price references can rarely be found, thus it can be a great challenge for sellers to set reasonable prices for their second-hand listings. In other words, the sellers will benefit a lot if the online second-hand platform, using their big data of various selling/sold items, can provide valuable price suggestions for their listings. Given the fact that online buyers rely heavily on the item images and text descriptions to make purchase decisions, we can make a safe assumption that the price of a second-hand item is mainly determined by its image and text description. Accordingly, in this work, we aim at designing a price suggestion system for the online second-hand platforms, which will provide effective price suggestions for online second-hand items with the uploaded images and texts.

However, for this real-word problem, there are many challenges we should keep in mind:

\textbf{Improvement of user engagement:} The success of a commercial application is rooted in the number of people using this application actively and the benefit it brings to the users. On one hand, the suggested prices provided by the designed system will maximize the gain of the sellers. On the other hand, the price suggestions should be able to promote transactions effectively.

\begin{figure}
\includegraphics[width=\linewidth]{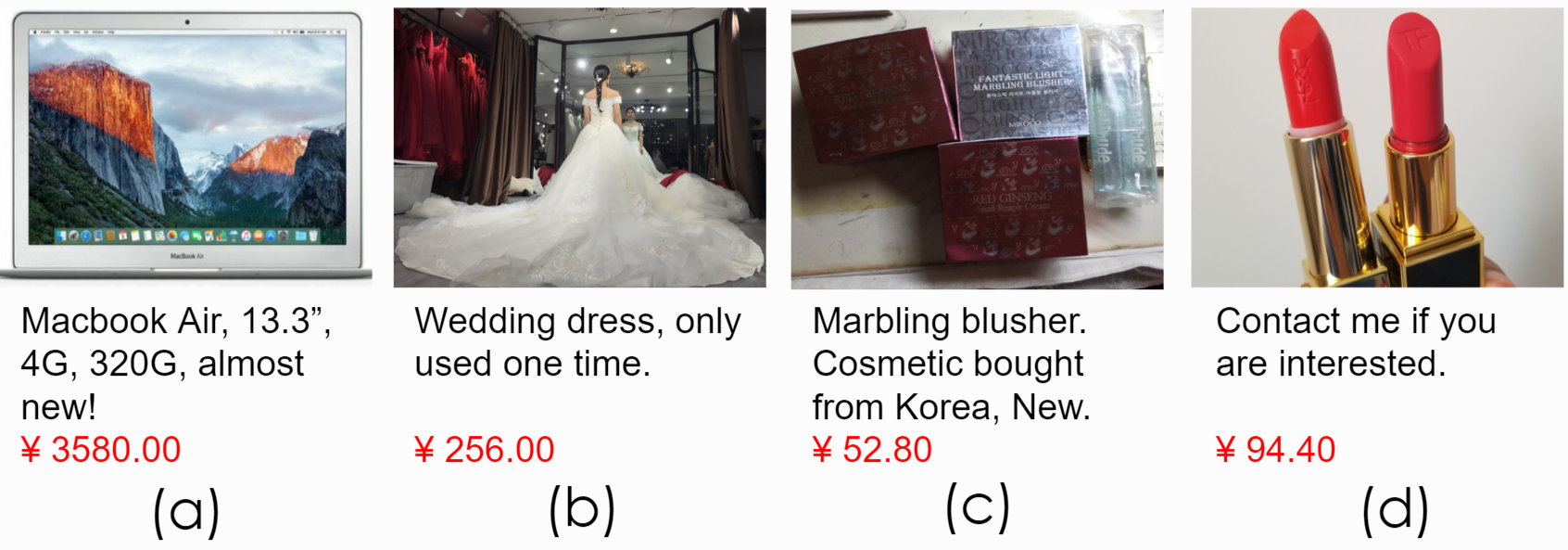}
\caption{Challenges for online second-hand item price suggestion. (a) text is more important, (b) image is more important (c) bad image quality, (d) useless text information.}
\label{fig:challenge}
\end{figure}

\textbf{Different importance of image and text:} Generally, the uploaded image and the text description of a second-hand item are complementary. Taking Figure~\ref{fig:challenge}(a) as an example, the text description gives specification information of the laptop, such as the model, memory, storage, etc. While the image offers a visual perception of this laptop's condition. Undoubtedly, we will rely more on the text information than the image information to estimate the price of this laptop, which demonstrates that image and text have different importance for price prediction of the second-hand items.

\textbf{Nonstandard image or text description:} Unlike the professional merchant, second-hand item sellers usually spend very limited time and effort on taking photos and writing text descriptions for listing the items. Accordingly, the quality of the images and text descriptions cannot be guaranteed. The uploaded images may have very bad image qualities (Figure~\ref{fig:challenge}(c)), and the text descriptions may only cover some useless information (Figure~\ref{fig:challenge}(d)).

Though the above challenges make the price suggestion problem very difficult, we still have some strengths to attack this challenging problem: (a) every online listed second-hand item has its own image and text description, with which we can determine whether it is feasible to predict the item's price with its image and text, and provide the price suggestion if it is feasible. Otherwise, the users are encouraged to update the images and texts; (b) some statistical features \footnote{Statistical features about the item price. We use the following statistical features in this work: the first, second, third quartile and the mean of the sold price of all sold second-hand items and the listing price of all listed but unsold second-hand items; the first, second, third quartile and the mean of the sold price of all sold second-hand items in each category and the listing price of all listed but unsold second-hand items in each category.} from the online second-hand platform can be leveraged to assist the price suggestion; and (c) benefiting from the increasing prevalence of online second-hand trading, we are able to collect tons of data over time to train and keep refining our price suggestion system.

Considering these challenges and strengths, we design a multi-modal price suggestion system which consists of a binary classification model and a regression model for online second-hand items. Specifically, first we determine whether the uploaded image and text description of a second-hand item along with some statistical features are qualified for price suggestion with the binary classification model; then, we predict prices for those whose images and texts are qualified with the regression model.

Our main contributions are four-fold:
\begin{itemize}
    \item We develop an intelligent price suggestion system for online second-hand items based on the uploaded images and text descriptions, which consists of a binary classification model to first determine whether the uploaded image and text are qualified to perform price suggestion for an item, and a regression model to provide price suggestions for items with qualified images and texts.
    \item Two different strategies are proposed to jointly train the classification model and the regression model, which can improve the performance of the price suggestion system. 
    \item A customized loss function is designed for optimizing the regression model to provide effective and reasonable price suggestions, which tries to not only maximize the gain of the sellers but also promote transactions.
    \item A set of metrics is designed to evaluate the price prediction performance of the system more effectively. 
\end{itemize}

\section{Related Work}
\label{sec:rework}

Price suggestion for items is to provide a price which is supposed to be accepted by both the seller and the buyer with a high probability. A large literature has explored various pricing strategies.

\begin{figure*}
\includegraphics[width=\linewidth]{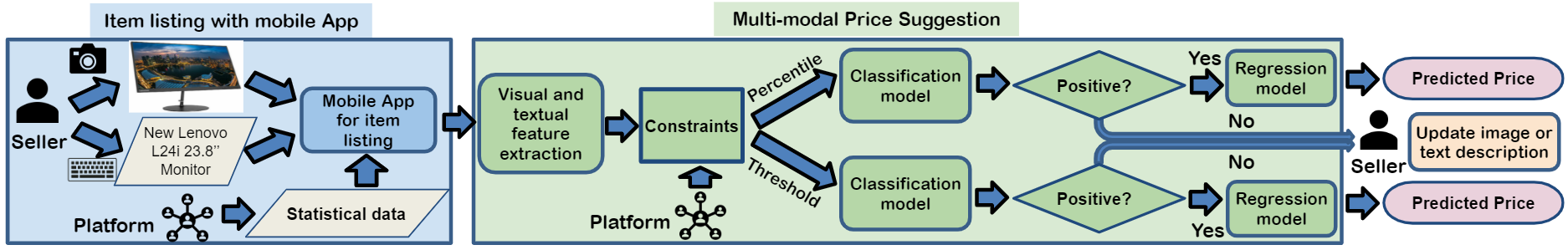}
\caption{The proposed multi-modal price suggestion system.}
\label{fig:flowchart}
\end{figure*}

Pudaruth ~\cite{pudaruth2014} predicts prices for second-hand cars with four different machine learning algorithms, namely k-nearest neighbor (KNN), Naive Bayes, linear regression, and decision trees, among which KNN performs the best. Raykhel and Ventura ~\cite{raykhel2009} develop a system which implements a feature-weighted KNN algorithm for attribute-based price prediction of eBay online auction trading, focusing on the laptop only. A heuristic approach to compute dynamic pricing strategies under competition is proposed in ~\cite{schlosser2018} for online marketplaces, the performance of which is evaluated in a real-life experiment on Amazon for the sale of second-hand books. Sun et al. ~\cite{sun2017} introduce a back propagation (BP) neural network algorithm to evaluate the price of the online second-hand cars. You et al. ~\cite{you2017} propose to use a recurrent neural network for online real estate price estimation, based on both the visual information and the location information of the real estate. Ye et al. ~\cite{ye2018} present a dynamic pricing strategy, in which a regression model trained with a customized loss function is used to estimate the price for shared houses at Airbnb. Law et al. ~\cite{law2019take} propose to estimate the price of a house by using the features extracted from the street view and satellite images of this house, in which way the surrounding environment is considered when estimating the house price. Maestre et al. ~\cite{maestre2018} show how to solve the unfair dynamic pricing problem with reinforcement learning by integrating fairness metrics as part of the model optimization, so that an effective balance between revenue and fairness maximization can be achieved. These methods, either use some conventional machine learning algorithms or employ deep learning, perform well for a certain item category such as books, cars or houses, but in our scenarios, we are targeting on provide price prediction for online second-hand items which fall into numerous fine-grained categories.

Recently, some efforts were made to provide price suggestions to online sellers by estimating reasonable prices for products they are selling, which have very similar goal with our work. A price suggestion challenge is held on Kaggle \footnote{https://www.kaggle.com/c/mercari-price-suggestion-challenge} by a Japanese E-commerce company, Mercari, which targets on giving price suggestions for online second-hand items based on their text descriptions. Han et al. \cite{han2019} proposed a vision-based price suggestion model to predict prices for online second-hand items. However, either only textual feature or only visual feature was involved in these works.

\section{Multi-modal Price Suggestion}
\label{sec:method}

The multi-modal price suggestion system is proposed to provide effective price suggestions for sellers to list their second-hand items online based on the visual and textual features extracted from the uploaded images and text descriptions, along with some statistical features. The overall workflow of the proposed price suggestion system is presented in Figure~\ref{fig:flowchart}. When a seller takes an image and writes a text description for a second-hand item and uploads to the mobile App to list this item, the uploaded image, text description and some statistical features obtained from the platform will be collected by the mobile App. Then, visual and textual features will be extracted from the image and text description, respectively. The platform operator can set different constraints on second-hand item price suggestion: (1) percentile constraint (i.e., how many percents of second-hand items on the platform are with qualified images and texts for reasonable price suggestion), and (2) threshold constraint (i.e., the second-hand item whose image and text are qualified for reasonable price suggestion should have its price prediction error less than a given threshold). With these two different constraints, two price suggestion modules are designed: (1) price suggestion with percentile constraint and (2) price suggestion with threshold constraint. Each module consists of a binary classification model taking as input the extracted visual and textual features along with some statistical features to determine whether the uploaded image and text are qualified for reasonable price suggestion, and a regression model to provide a suggested price for an item with qualified image and text description. For an item with unqualified image and text description, the user will be encouraged to update the item image and/or text description for better price suggestion. Note that the binary classification model plays an important role in our price suggestion framework. First, items with unqualified image and text are quite challenging to predict prices, and an unreasonable price suggestion will impress the sellers badly and thus hurt the user engagement. Thankfully, the binary classification model can filter out these items. Second, for the items filtered out by the binary classification model, sellers are encouraged to update their images and text descriptions, with which these items are more likely to be sold, and thus the transaction is promoted.


\subsection{Data Collection and Analysis}
\label{sec:Data}

\begin{figure}
\includegraphics[width=\linewidth]{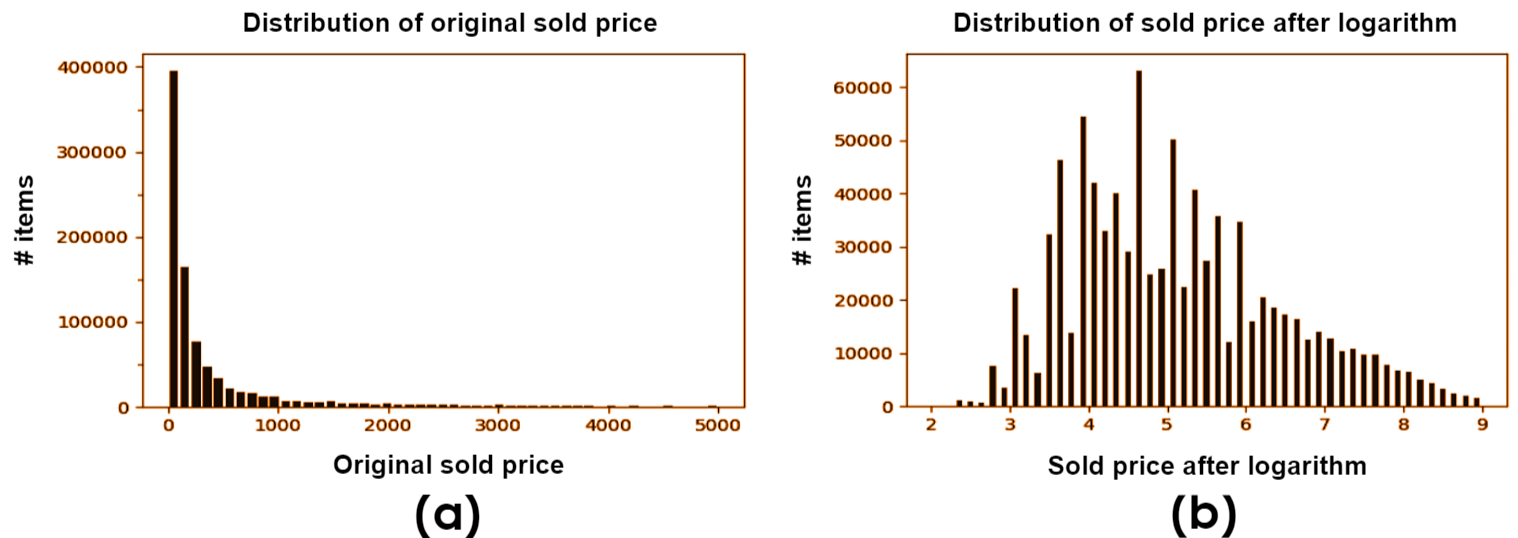}
\caption{Distribution of the original sold price (a) and the sold price after the logarithm operation (b).}
\label{fig:price}
\end{figure}

We collected a real-world dataset which contains more than 5 million online second-hand items over three months to develop our price suggestion system.
Unfortunately, the price distribution of these online second-hand items is heavily long-tailed (Figure~\ref{fig:price}(a)). Compared to long-tailed distribution, a Gaussian distribution is much easier to model. Thus, the logarithm operation is performed on the original second-hand item prices to transform the long-tailed distribution to a Gaussian-like distribution (Figure~\ref{fig:price}(b)).

\subsection{Multi-modal Feature Extraction}
\label{sec:Feature}

To develop a multi-modal price prediction model, we first need to extract effective visual and textual features from the image and text. With the prevalence of deep learning, various networks have been designed to extract features from images, such as VGG~\cite{simonyan2014}, AlexNet~\cite{krizhevsky2012} and ResNet~\cite{he2016deep}, etc. In this work, we extract the representative visual feature from the uploaded item image with a pre-trained CNN model proposed in ~\cite{zhang2018}. This CNN model is able to detect the item from an uploaded image with complex background and extract visual features of the detected item. Moreover, this model is trained on a real-world dataset (some item images uploaded by Alibaba Pailitao users) that is quite similar to ours, which guarantees the effectiveness of this model on our task.

To extract the textual feature from the text description, we create a corpus that contains single words, each of which has a unique indicator number. We represent the text description by an indicator vector which consists of the indicator numbers of the single words in this text description (i.e., one-hot encoding). After statistically checking the distribution of the text length, we fix the indicator vector length as 32 (i.e., if a text description is shorter than 32, we complement its indicator vector with `0's, if a text description is longer than 32, we only take the first 32 indicator numbers), which is longer than the lengths of 90\% of the text descriptions. The feature embedding ~\cite{cheng2016} operation is implemented on the discrete indicator vector, which converts the sparse, high-dimensional indicator features into low-dimensional and dense real-valued features. Note the embedding vector is randomly initialized and then the values are updated to minimize the final loss function during training.

For some second-hand items, images shows more information of the items than their text descriptions, while for some others, the text descriptions are more important for presenting the items. Accordingly, instead of directly concatenating the extracted visual and textual features together for price prediction, we add an attention mechanism to the concatenation (as shown in the red dashed box in Figure~\ref{fig:model}). We add a fully connected layer after the concatenated feature vector to get a vector with two weights (visual weight and textual weight) for these two kinds of features, then a softmax operation is applied on these two weights. Finally, we multiply the visual feature with the visual weight, and the textual feature with the text weight to get the final visual and textual features.

\subsection{Design of Multi-modal Price Suggestion}
\label{sec:logic}

\begin{figure}
\includegraphics[width=\linewidth]{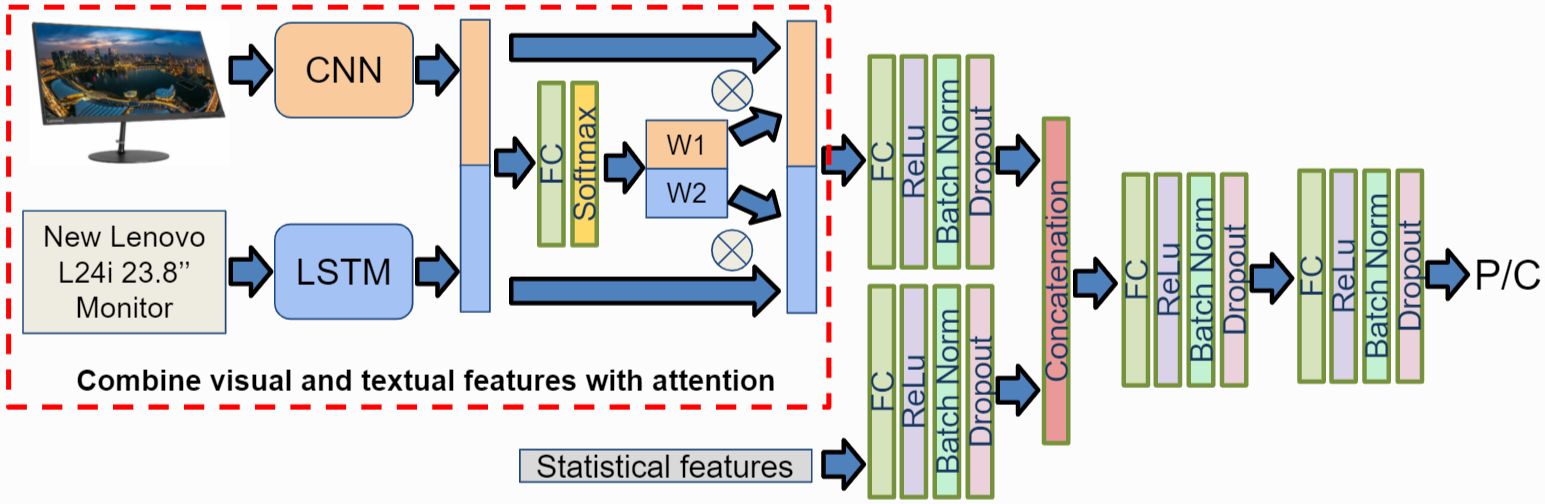}
\caption{Architecture of the regression/classification model. The output of the regression model is a predicted price $P$, while the classification model outputs a judgement $C$ of whether the extracted visual and textual features of the item are qualified for price prediction.}
\label{fig:model}
\end{figure}

With the extracted visual and textual features from the item image and text description, now we are expected to train a regression model to perform price prediction for online second-hand items. Unfortunately, as described in the challenges in Section~\ref{sec:intro}, some uploaded images and text descriptions are not qualified for offering reasonable price suggestions for the items. Thus, before predicting price for an online second-hand item, we first need a binary classification model to judge whether the uploaded image and text description of this item are qualified for offering a reasonable price suggestion, because a bad price suggestion can impress the seller very badly and thus hurt the user engagement. Accordingly, in the proposed multi-modal price suggestion system, a binary classification model will first judge whether the extracted visual and textual features of an item are qualified for reasonable price prediction, and then a regression model will perform price prediction for items with qualified visual and textual features. Figure~\ref{fig:model} presents the architecture of the binary classification model and the regression model, which share the same architecture, except for the last layer. However, no parameter is shared by these two models. The model takes the extracted visual and textual features of the item along with some statistical features as input, and outputs a judgement of whether the extracted visual and textual features of the item are qualified for reasonable price prediction (classification model), or a predicted price for the second-hand item (regression model).

To train the binary classification model, some positive samples (items whose images and texts are qualified for reasonable price suggestions) and negative samples (items whose images and texts are not qualified for reasonable price suggestions) are expected to be labeled. However, \emph{the labels of the positive and the negative samples are determined by the price prediction regression model} (items with small price prediction errors should be labeled as positive, while items with large price prediction errors should be labeled as negative). \emph{In turn, the performance of the regression model will be greatly influenced by the binary classification model}, as it provides price suggestions only for positive items classified by the classification model. \emph{Thus, the regression and classification tasks are mingled together. Instead of training the classification model and regression model separately, we consider to train them in a joint way.}

\subsubsection{Joint Optimization with Percentile Constraint}
\label{sec:percentile}
There are various criteria to label training samples, i.e., whether the image and text description of an item are qualified for offering a reasonable price suggestion (positive) or not (negative). The operators of the online trading platforms can adopt a percentile strategy based on their statistical data, e.g., 50\% of the items are positive, and the other 50\% are negative. In this scenario, the classification model should regard the items with the top 50\% best price suggestions as positive, and the others as negative. The loss function for jointly training the classification model and the regression model with the percentile constraint is:

\begin{equation}
\begin{split}
    \min_{\Theta_{1},\Theta_{2}} \Big\{&\frac{1}{N}\sum_{i=1}^{N}f_{cls}(x_{i};\Theta_{1}) \cdot L\Big(y_{i}, f_{reg}(x_{i};\Theta_{2})\Big) \\ &+ \beta \cdot max\Big(0, \delta-\frac{1}{N}\sum_{i=1}^{N}I(f_{cls}(x_{i};\Theta_{1}))\Big) \Big\},
\end{split}
\label{Eq:percent}
\end{equation}
where $f_{cls}$ and $f_{reg}$ represent the binary classification model and the price prediction regression model in the price suggestion system, respectively, whose architectures are shown in Figure~\ref{fig:model}; $\Theta_{1}$ and $\Theta_{2}$ are the parameters of $f_{cls}$ and $f_{reg}$, respectively; 
$N$ denotes the number of training samples; $x_{i}$ is the $i^{th}$ training sample; $L$ is the loss function used for optimizing the regression model, which will be introduced in details in Section ~\ref{sec:loss}; $\delta$ is a prior to control the percentage of positive items (items with qualified images and text descriptions for reasonable price suggestion), i.e., how many percents of items can be regarded as positive; and $\beta$ is a weight to balance the two terms in Eq.~\ref{Eq:percent}. For the output of the classification model $f_{cls}(x_{i};\Theta_{1})$ in the first term, rather than use binary classification variables, we instead adopt real-valued classification variables which are in the range from 0 to 1. The advantages are two-fold: first, the continuous real value can help us simultaneously optimize the binary classification model $f_{cls}$ and the regression model $f_{reg}$ in an easy way; second, by relaxing the binary output of the classification model to a continuous variable from 0 to 1, the confidence of an item being positive can be captured. $I(\cdot)$ is an indicator function to convert the output of $f_{cls}(x_{i};\Theta_{1})$ from continuous values $[0,1]$ to discrete values $\{0,1\}$:
\begin{equation}
    I(f_{cls}(x_{i};\Theta_{1})) = \begin{cases}
    0& \text{if $f_{cls}(x_{i};\Theta_{1}) < 0.5$}\\
    1& \text{otherwise},
\end{cases}
\label{Eq:C1}
\end{equation}

The first term in Eq.\ref{Eq:percent} tries to achieve two goals: first, optimizing the price regression model $f_{reg}$ by minimizing a specific designed loss function $L$ mainly with the positive samples labeled by the binary classification model $f_{cls}$; second, optimizing the binary classification model $f_{cls}$ by enforcing it to label the samples with the most reasonable price suggestions (samples with the smallest loss $L(y_{i}, f_{reg}(x_{i};\Theta_{2}))$) as positive. The second term in Eq.\ref{Eq:percent} is a prior constraint on at least how many percents of items should be regarded as positive. This term also helps avoid a trivial solution that the binary classification model labels all the samples as negative.

\subsubsection{Joint Optimization with Threshold Constraint}
\label{sec:threshold}
There is a prior parameter $\delta$ in the percentile-based joint optimization, which requires us to have some prior knowledge of the quality of the item images and texts to set a suitable percentile value. When the operator of the second-hand trading platform is not sure about this prior knowledge, an alternative criterion for labeling the samples to train the binary classification model is to set a threshold $\epsilon$ for the price prediction accuracy, i.e., if the loss $L(y_{i}, f_{reg}(x_{i};\Theta_{2}))$ of an item is smaller than $\epsilon$, this item is regarded as a positive training sample, otherwise, it is negative. In this scenario, it is highly possible that the positive and negative labels are imbalanced. To relieve the impact of imbalanced labeling, we adopt a weighted cross entropy loss for the binary classification model ~\cite{yan2019}, and the final loss function for the joint optimization of the regression model and the classification model with threshold constraint is
\begin{small}
\begin{equation}
\begin{split}
    \min_{\Theta_{1},\Theta_{2}} \! \frac{1}{N}\sum_{i=1}^{N}\!\Big\{f_{cls}&(x_{i};\Theta_{1}) \cdot L\Big(y_{i}, f_{reg}(x_{i};\Theta_{2})\Big) \\
    - \gamma \!\cdot\! \!\Big[&w_{p} \!\cdot\! C(x_{i},y_{i};\Theta_{2}) \cdot log\big(f_{cls}(x_{i}; \Theta_{1})\big) \\
    &\! w_{n} \!\cdot\!\Big(1\!-\!C(x_{i},y_{i};\Theta_{2})\Big) \!\cdot\! log\big(1\!-\!f_{cls}(x_{i}; \Theta_{1})\!\big)\!\Big]
    \Big\},
\end{split}
\label{Eq:threshold}
\end{equation}
\end{small}
where $\gamma$ is a weight to scale the cross-entropy loss term, $w_{p}$ and $w_{n}$ are the weights for weighted cross entropy loss, which are defined as $w_{p}=\sqrt{(P+N)/2P}$, $w_{n}=\sqrt{(P+N)/2N}$, where $P$ and $N$ are the numbers of positive and negative training samples in a training batch, respectively. $C$ is an indicator function defined as 
\begin{equation}
    C(x_{i}, y_{i};\Theta_{2}) = \begin{cases}
    0& \text{if $L\Big(y_{i}, f_{reg}(x_{i};\Theta_{2})\Big) > \epsilon$}\\
    1& \text{otherwise}
\end{cases}.
\label{Eq:epsilon}
\end{equation}

The first term in Eq.\ref{Eq:threshold} and the first term in Eq.\ref{Eq:percent} play similar roles in these two equations. While the second term in Eq.\ref{Eq:threshold} is a weighted cross entropy loss to optimize the binary classification model with the training samples labeled by the regression model under the threshold constraint.

\subsection{Loss Function for Regression Model}
\label{sec:loss}
To train the regression model in the joint optimization strategy, we need to design an objective function for the regression model (i.e., the loss function $L$ in Eq.\ref{Eq:percent} and Eq.\ref{Eq:threshold}). As mentioned in the challenges in Section~\ref{sec:intro}, to improve the user engagement, the system should be able to provide price suggestions for second-hand items that achieves two goals: maximizing the gain for the seller and promoting transactions. Thus, instead of adopting a conventional loss function such as mean square loss, we should design a customized objective function for the regression model in our price suggestion system. As there are two statuses for a listed second-hand item, sold and unsold, we will consider the following two cases separately:

\begin{itemize}
    \item If a listed second-hand item is sold at price $p^{sold}$, the suggested price $p^{sug}$ for this item should not be lower than the sold price $p^{sold}$. Otherwise, if we offer a lower suggested price $p^{sug}$ for this item and the seller adopts our suggested price, the seller will lose a revenue of $p^{sold}-p^{sug}$. However, the suggested price $p^{sug}$ should not be too high than the sold price $p^{sold}$ to harm the transaction.
    
    \item If a listed second-hand item is not sold, this means the listing price $p^{list}$ of this item is too high for transaction. Thus, the suggested price $p^{sug}$ should be lower than the listing price $p^{list}$ to promote the transaction. But the suggested price $p^{sug}$ can not be too lower than the listing price $p^{list}$ to decrease the revenue of the seller.
\end{itemize}

Inspired by the above motivations, we develop a customized loss function $L$ for the regression model as follows:
\begin{equation}
    L\Big(y_{i}, f_{reg}(x_{i};\Theta_{2})\Big) = \begin{cases}
    U(p^{sug}, p^{sold}) & \text{if sold}\\
    V(p^{sug}, p^{list}) & \text{if unsold}
\end{cases},
\label{Eq:L}
\end{equation}
where
\begin{small}
\begin{equation}
    U(p^{sug}, p^{sold}) = \begin{cases}
    p^{sold} \!-\! p^{sug} & \text{if $p^{sug}\!<\!p^{sold}$} \\
    0 & \text{if $p^{sold} \!\leq \! p^{sug} \!\leq \! \mu \!\cdot\! p^{sold}$} \\
    p^{sug} \!-\! \mu \!\cdot\! p^{sold} & \text {if $p^{sug} \!>\! \mu \!\cdot\! p^{sold}$}
\end{cases},
\label{Eq:U}
\end{equation}
\end{small}
and
\begin{small}
\begin{equation}
    V(p^{sug}, p^{list}) = \begin{cases}
    \nu \cdot p^{list} \!-\! p^{sug} & \text{if $p^{sug}\!<\! \nu \cdot p^{list}$} \\
    0 & \text{if $\nu \cdot p^{list} \!\leq \! p^{sug} \!\leq \! p^{list}$} \\
    p^{sug} \!-\! p^{list} & \text {if $p^{sug} \!>\! p^{list}$}
\end{cases},
\label{Eq:V}
\end{equation}
\end{small}
where $\mu$ and $\nu$ are two parameters controlling the make-up rate of the sold price and the discount rate of the listing price, respectively. $\mu=1.2$ and $\nu=1/1.2$ in our experiments (by analyzing all the sold items, we find the original listing prices are about 1.2 times of the final sold prices). It is worth noting that all the prices here are log prices (i.e., applying the logarithm operation on all the original prices), and the original price unit is the Chinese Yuan (CHN).

From the designed loss function (Eq.\ref{Eq:L} along with Eq.\ref{Eq:U} and Eq.\ref{Eq:V}) we can see that, there does not exist an ground-truth optimal price to guide the predicted price for a second-hand item, but we have a \textbf{\emph{target price range}} $[p^{sold}, \mu \!\cdot \!p^{sold}$] for the sold items and another \textbf{\emph{target price range}} $[\nu \!\cdot \!p^{list}, p^{list}]$ for unsold items, and the predicted prices are expected to fall in the target price ranges.

In all of our experiments, the classification model and the regression model are jointly trained with a fixed learning rate of 0.0005 for 3,400 epochs, and then with a fixed learning rate of 0.0002 for 1,700 more epochs. The batch size is 4096, and the Adam optimizer ~\cite{kingma2014} is adopted for optimization.

\section{Experiments}
We conduct exclusive experiments to evaluate the performance of the proposed price suggestion system. First, the collected dataset and evaluation metrics used in our experiments are introduced (Section~\ref{sec:data}-~\ref{sec:metric}). Second, we evaluate the performance of the multi-modal price suggestion quantitatively (Section~\ref{sec:percenteval}-~\ref{sec:baseline}). Third, we perform analysis on the features used in our system (Section~\ref{sec:feature}).

\subsection{Dataset}
\label{sec:data}
A large real-world dataset is collected from Xianyu, an online second-hand platform launched by China's E-commerce giant, Alibaba, to evaluate the developed system. The collected data cover various item classes, such as clothes, shoes, cosmetic, phone, etc., which are divided into three sets: training set, validation set, and testing set. Validation set is used to determine the hyper-parameters $\beta$ in Eq.~\ref{Eq:percent} and $\gamma$ in Eq.~\ref{Eq:threshold}. Table~\ref{tab:data} summarizes the data used in our experiments.

\begin{table}
\begin{center}
\caption{Summary of dataset used in our experiments.} \label{tab:data}
\resizebox{\columnwidth}{!}{
\begin{tabular}{|c|c|c|c|}
  \hline
   & \# sold items & \# unsold items & total \\
  \hline
  \hline
  training set & 2,937,851 & 1,385,756 & 4,323,607 \\
  \hline
  validation set & 150,916 & 70,785 & 221,701  \\
  \hline
  testing set & 671,774 & 326,257 & 998,031  \\
  \hline
\end{tabular}
}
\end{center}
\end{table}

\subsection{Evaluation Metric}
\label{sec:metric}

As discussed in Section ~\ref{sec:loss}, \emph{we do not have a ground-truth optimal price for the second-hand item which can not only maximize the gain for the seller but also promote the transaction. Therefore, we derive a set of metrics based on the motivations discussed in Section~\ref{sec:loss} to evaluate the goals of our price suggestion system}:

\begin{itemize}
    \item \textbf{Sold Mean Log Error (SMLE)}: how close the suggested prices are to the target price range for all sold items (the number of all those sold items is $I_{1}$). \emph{This metric measures the overall price prediction accuracy for all sold items.}
    \begin{small}
    \begin{equation}
        SMLE = \frac{1}{I_{1}}\sum_{i=1}^{I_{1}}\Big(\min(p_{i}^{sold}\!-\!p_{i}^{sug}, \ p_{i}^{sug}\!-\!\mu \!\cdot\! p_{i}^{sold}, \ 0)\Big)
    \label{Eq:SMLE}
    \end{equation}
    \end{small}
    
    \item \textbf{Sold Price Decrease Mean Log Error (SPDMLE)}: for sold items whose suggested prices are lower than the lower boundary of the target price range (the number of those items is $I_{2}$), the distance between the suggested prices and the lower boundary. \emph{This metric measures the potential economic losses of the sellers if the second-hand items are sold with the predicted prices, i.e., if this metric is lower, the economic loss is reduced for sellers.}
    \begin{small}
    \begin{equation}
        SPDMLE = \frac{1}{I_{2}}\sum_{i=1}^{I_{2}}(p_{i}^{sold}\!-\!p_{i}^{sug})
    \label{Eq:SPDMLE}
    \end{equation}
    \end{small}
    
    \item \textbf{Sold Price Increase Mean Log Error (SPIMLE)}: for sold items whose suggested prices are higher than the higher boundary of the target price range (the number of those items is $I_{3}$), the distance between the suggested prices and the higher boundary. \emph{This metric reflects the potential harm to the transactions if items are set with the predicted prices, i.e., if this metric is lower, the transactions are promoted.}
    \begin{small}
    \begin{equation}
        SPIMLE = \frac{1}{I_{3}}\sum_{i=1}^{I_{3}}(p_{i}^{sug}\!-\!\mu \!\cdot\! p_{i}^{sold})
    \label{Eq:SPIMLE}
    \end{equation}
    \end{small}
    
    \item \textbf{Unsold Mean Log Error (UMLE)}: how close the suggested prices are to the target price range for all listed but unsold items (the number of all those listed but unsold items is $I_{4}$). \emph{This metric measures the overall price prediction accuracy for all unsold items.}
    \begin{small}
    \begin{equation}
        UMLE = \frac{1}{I_{4}}\sum_{i=1}^{I_{4}}\Big(\min(\nu \!\cdot\! p_{i}^{list}\!-\!p_{i}^{sug}, \ p_{i}^{sug}\!-\!p_{i}^{list}, \ 0)\Big)
    \label{Eq:UMLE}
    \end{equation}
    \end{small}
    
    \item \textbf{Unsold Price Decrease Mean Log Error (UPDMLE)}: for unsold items whose suggested prices are lower than the lower boundary of the target price range (the number of those items is $I_{5}$), the distance between the suggested prices and the lower boundary. \emph{This metric reflects the potential economic loss if items are sold with the predicted prices, i.e., if this metric is lower, the potential economic loss is reduced for sellers if items are sold with the predicted prices.}
    \begin{small}
    \begin{equation}
        UPDMLE = \frac{1}{I_{5}}\sum_{i=1}^{I_{5}}(\nu \!\cdot\! p_{i}^{list}\!-\!p_{i}^{sug})
    \label{Eq:UPDMLE}
    \end{equation}
    \end{small}
    
    \item \textbf{Unsold Price Increase Mean Log Error (UPIMLE)}: for unsold items whose suggested prices are higher than the higher boundary of the target price range (the number of those items is $I_{6}$), the distance between the suggested prices and the higher boundary. \emph{This metric reflects the potential harm to the transactions if items are set with the predicted prices, i.e., if this metric is lower, the potential transactions are promoted.}
    \begin{small}
    \begin{equation}
        UPIMLE = \frac{1}{I_{6}}\sum_{i=1}^{I_{6}}(p_{i}^{sug}\!-\!p_{i}^{list})
    \label{Eq:UPIMLE}
    \end{equation}
    \end{small}
\end{itemize}

In the above evaluation metrics, in addition to computing the error with original prices, we can compute all the errors with log prices. The errors computed with log price are relative error ($log(p_{1})-log(p_{2}) = log(\frac{p_{1}}{p_{2}})$), which is more reasonable than the absolute error in our scenario (e.g., predicting a \$1000 item as \$995 is supposed to be more accurate than predicting a \$10 item as \$5, though both absolute errors are \$5).

\subsection{Percentile-based Price Suggestion Evaluation}
\label{sec:percenteval}
First we evaluate the performance of the multi-modal price suggestion with the percentile constraint. The classification model ($f_{cls}$ in Eq.~\ref{Eq:percent}) and the regression model ($f_{reg}$ in Eq.~\ref{Eq:percent}) are jointly trained with the percentile constraint (i.e., how many percents of items with qualified images and texts for multi-modal price suggestion). When testing, the classification model first distinguishes the positive items (items with qualified images and texts for multi-modal price suggestion) from the negative ones (items with unqualified images and texts for multi-modal price suggestion), then the regression model predicts prices for the positive items. The results are summarized in Table~\ref{tab:percent}, from which we can see: (1) the percentage of the positive items classified by the classification model almost exactly equals to the given percentile constraint (e.g., given percentile constraint as 40\%, the classification model regards 40.05\% of the items as positive), which demonstrates the effectiveness of the classification model; (2) under all percentile constraints, the metric values of the positive samples classified by the classification model are much lower than the ones of the negative samples, which illustrates the effectiveness of both the classification model and the regression model; (3) as the percentage of positive items increases (i.e., more items are classified as positive by the classification model), the prediction errors of the positive items increase. This is intuitive since the quality control for images and texts is worsen with the increase of the percentage of positive items.

\begin{table}
  \caption{Evaluation of multi-modal price prediction with the percentile constraint. Positive items are those with qualified images and texts for multi-modal price suggestion.}
  \label{tab:percent}
  \resizebox{\columnwidth}{!}{
  \begin{tabular}{|c|c|c|c|c|c|c|}
    \hline
    & percentile & 40 & 50 & 60 & 70 & 80 \\
    \hline
    \hline
    & \# positive items & 399705 & 525232 & 617840 & 711960 & 812258 \\
    & \% positive items & 40.05\% & 52.63\% & 61.91\% & 71.34\% & 81.39\% \\
    \hline
    \multirow{6}{*}{positive} & SMLE & 0.1447 & 0.1466 & 0.1498 & 0.1579 & 0.1713 \\
    \cline{3-7}
               ~ & SPDMLE & 0.1798 & 0.1777 & 0.1810 & 0.1885 & 0.2011 \\
    \cline{3-7}
               ~ & SPIMLE & 0.1888 & 0.1971 & 0.2007 & 0.2087 & 0.2201 \\
    \cline{3-7}
               ~ & UMLE & 0.1096 & 0.1101 & 0.1107 & 0.1177 & 0.1291 \\
    \cline{3-7}
               ~ & UPDMLE & 0.1445 & 0.1429 & 0.1466 & 0.1568 & 0.1695 \\
    \cline{3-7}
               ~ & UPIMLE & 0.1504 & 0.1539 & 0.1547 & 0.1611 & 0.1694 \\
    \hline
    \multirow{6}{*}{negative} & SMLE & 0.5208 & 0.4482 & 0.4148 & 0.3946 & 0.4130 \\
    \cline{3-7}
               ~ & SPDMLE & 0.7002 & 0.5960 & 0.5213 & 0.4800 & 0.4860 \\
    \cline{3-7}
               ~ & SPIMLE & 0.3056 & 0.3466 & 0.3868 & 0.3890 & 0.4169 \\
    \cline{3-7}
               ~ & UMLE & 0.6552 & 0.5137 & 0.4411 & 0.3940 & 0.3907 \\
    \cline{3-7}
               ~ & UPDMLE & 0.8073 & 0.6651 & 0.5724 & 0.5106 & 0.4873 \\
    \cline{3-7}
               ~ & UPIMLE & 0.2284 & 0.2588 & 0.3010 & 0.2991 & 0.3256 \\
    \hline
\end{tabular}}
\end{table}

\begin{table}
  \caption{Quantitative evaluation of the regression model for multi-modal price prediction under different threshold constraints.}
  \label{tab:threshold}
  \resizebox{\columnwidth}{!}{
  \begin{tabular}{|c|c|c|c|c|c|c|}
    \hline
    & threshold & 0.100 & 0.125 & 0.150 & 0.175 & 0.200 \\
    \hline
    \hline
    & \# positive items & 436925 & 524502 & 592534 & 661807 & 722761 \\
    & \% positive items & 43.78\% & 52.55\% & 59.37\% & 66.31\% & 72.42\%\\
    \hline
    \multirow{6}{*}{positive} & SMLE & 0.1198 & 0.1266 & 0.1340 & 0.1472 & 0.1477 \\
    \cline{3-7}
               ~ & SPDMLE & 0.1449 & 0.1554 & 0.1636 & 0.1739 & 0.1770 \\
    \cline{3-7}
               ~ & SPIMLE & 0.1776 & 0.1830 & 0.1886 & 0.1996 & 0.2017 \\
    \cline{3-7}
               ~ & UMLE & 0.0833 & 0.0884 & 0.0942 & 0.1042 & 0.1065 \\
    \cline{3-7}
               ~ & UPDMLE & 0.1052 & 0.1190 & 0.1268 & 0.1353 & 0.1437 \\
    \cline{3-7}
               ~ & UPIMLE & 0.1353 & 0.1363 & 0.1421 & 0.1513 & 0.1522 \\
    \hline
    \multirow{6}{*}{negative} & SMLE & 0.3161 & 0.3219 & 0.3463 & 0.3373 & 0.3620 \\
    \cline{3-7}
               ~ & SPDMLE & 0.3891 & 0.3958 & 0.4393 & 0.4016 & 0.4006 \\
    \cline{3-7}
               ~ & SPIMLE & 0.3342 & 0.3337 & 0.3420 & 0.3514 & 0.4030 \\
    \cline{3-7}
               ~ & UMLE & 0.2775 & 0.2873 & 0.3106 & 0.3077 & 0.3213 \\
    \cline{3-7}
               ~ & UPDMLE & 0.3711 & 0.3800 & 0.4116 & 0.3942 & 0.3789 \\
    \cline{3-7}
               ~ & UPIMLE & 0.2635 & 0.2625 & 0.2699 & 0.2739 & 0.3386 \\
    \hline
\end{tabular}}
\end{table}

\begin{table*}
  \caption{Quantitative comparison with baseline and ablation study. The metrics are computed on positive items only.}
  \label{tab:baseline}
  \begin{tabular}{|c|c|c|c|c|c|c|c|c|}
    \hline
     constraint & method & \# positive items & SMLE & SPDMLE & SPIMLE & UMLE & UPDMLE & UPIMLE \\
    \hline
    \hline
    \multirow{5}{*}{percentile} & Baseline & 617840 & 0.1514 & 0.1881 & 0.2007 & 0.1140 & 0.1558 & 0.1547 \\
    \cline{2-9}
               ~ & Ours & 617840 & 0.1498 & 0.1810 & 0.1957 & 0.1107 & 0.1466 & 0.1526 \\
    \cline{2-9}
               ~ & Ours W/O attention & 614021 & 0.1597 & 0.1857 & 0.2096 & 0.1153 & 0.1450 & 0.1641 \\
    \cline{2-9}
               ~ & Ours W/O image & 612013 & 0.1604 & 0.1839 & 0.2134 & 0.1162 & 0.1452 & 0.1660 \\
    \cline{2-9}
               ~ & Ours W/O text & 617740 & 0.2157 & 0.2460 & 0.2614 & 0.1768 & 0.2185 & 0.1478 \\
    \hline
    \multirow{5}{*}{threshold} & Baseline & 592534 & 0.1400 & 0.1722 & 0.1877 & 0.1015 & 0.1318 & 0.1478 \\
    \cline{2-9}
               ~ & Ours & 592534 & 0.1340 & 0.1636 & 0.1886 & 0.0942 & 0.1268 & 0.1421 \\
    \cline{2-9}
               ~ & Ours W/O attention & 516231 & 0.1495 & 0.1743 & 0.2024 & 0.1099 & 0.1373 & 0.1586 \\
    \cline{2-9}
               ~ & Ours W/O image & 518857 & 0.1505 & 0.1746 & 0.2062 & 0.1073 & 0.1318 & 0.1594 \\
    \cline{2-9}
               ~ & Ours W/O text & 423097 & 0.1733 & 0.2015 & 0.2304 & 0.1364 & 0.1686 & 0.1870 \\
    \hline
\end{tabular}
\end{table*}

\subsection{Threshold-based Price Suggestion Evaluation}
\label{sec:thresholdeval}
We then test the multi-modal price suggestion with the threshold constraint. The classification model and the regression model are jointly trained with Eq.~\ref{Eq:threshold}. Intuitively, the joint training and the performance of these two models are greatly influenced by the given threshold value $\epsilon$ in Eq.~\ref{Eq:epsilon}. Table~\ref{tab:threshold} summaries the performance of the regression model under different threshold constraints. We can get the following conclusions from the results: (1) when the given threshold value $\epsilon$ in Eq.~\ref{Eq:epsilon} increases, i.e., the restriction on the definition of positive items is loosen, the classification model adopts more second-hand items for price prediction by the regression model. In this scenario, the errors of the positive items classified by the classification model will increase (i.e., the overall performance of the regression model on positive items gets worse). Thus, when evaluating these two models, we should not only focus on the evaluation metrics of the positive items, but also keep an eye on the number of classified positive items; and (2) under different threshold constraints, the regression model performs much better on positive items (classified by the classification model) than on negative ones, which demonstrates the effectiveness of the regression model and the classification model. 

\subsection{Comparison with Baseline}
\label{sec:baseline}
For the multi-modal price prediction with percentile constraint (percent constraint on positive items is 60\%), we design a regression model which takes as input the multi-modal features (image feature, text feature and statistical features) and outputs predicted prices, and use it as a baseline model. Note the baseline regression model shares the same architecture with the regression model in our proposed system, but it is trained separately, not jointly trained with the classification model. Then we classify the items into positive and negative with the classification model in our proposed system, and perform price prediction for positive items with the baseline regression model and the jointly-trained regression model (the proposed one), respectively. We also implement the similar comparison for the multi-modal price prediction with threshold constraint (threshold value $\epsilon$ in Eq.~\ref{Eq:epsilon} is 0.15). The comparison results are presented in Table~\ref{tab:baseline} (the `Baseline' rows and the `Ours' rows), which shows that in both cases (percentile constraint and threshold constraint) the proposed regression model performs better than the baseline model on positive items. The reason is that in our system the regression model is jointly trained with the classification model, which means the proposed regression model is trained mainly with positive items classified by the classification model, while the baseline regression model is trained with all training items. The negative items can degrade the optimization of the baseline regression model and its performance on positive items.

Note that we use relative log error for performance evaluation, take the $SPDMLE$ (Sold Decrease Mean Log Error) as an example, if for the baseline, $SPDMLE = p_{sold} - p_{sug} = log(p^{orig}_{sold}) - log(p^{orig}_{sug}) = log(\frac{p^{orig}_{sold}}{p^{orig}_{sug}}) = 0.15 \Rightarrow \frac{p^{orig}_{sold}}{p^{orig}_{sug}} = e^{0.15} = 1.162 \Rightarrow p^{orig}_{sold} = 1.162 \ast p^{orig}_{sug}$, where $p^{orig}_{sold}$ and $p^{orig}_{sug}$ are the original sold and predicted prices before the logarithm operation, respectively. This means that the baseline achieves a $16.2\%$ absolute prediction mean error. Similarly, for the proposed system with $SPDMLE = 0.14$, we have $p^{orig}_{sold} = 1.150 \ast p^{orig}_{sug}$, and the proposed system achieves a $15.0\%$ absolute prediction mean error. The prediction improvement is $(16.2\%-15.0\%)/15.0\% = 8\%$. Thus, even the difference between metric values is small (0.01), the prediction accuracy improvement by the proposed price prediction system is significant.

To our best knowledge, this is the first work to predict prices for second-hand items with texts and images, and with the goal of maximizing the gain for the seller and promoting the transactions for the platform simultaneously. Thus, we do not conduct comparison between our model and other price prediction models with the proposed metrics, which we think is an unfair comparison.

\subsection{Ablation studies}
\label{sec:feature}
In this subsection, we perform ablation studies on visual and textual features to check what important roles they are playing on price prediction. First, we jointly train a classification model and a regression model under percentile constraint (percent constraint on positive items is 60\%) without visual feature, and evaluate the price prediction performance. Then, we jointly train these two models but without textual feature and evaluate the price prediction performance. After that, we compare the results of these two experiments with the results of our proposed method. We also do this for joint training with threshold constraint (threshold value $\epsilon$ in Eq.~\ref{Eq:epsilon} is 0.15). The comparison results are summarized in Table~\ref{tab:baseline} (`Ours W/O image' rows, `Ours W/O text' rows and `Ours' rows), which shows that both the visual feature and the textual features contribute to the price suggestion system, while the textual feature plays a much more important role on price prediction for second-hand items than the visual feature. The reason is that for most of the second-hand items, text descriptions provide much more important information than images. For example, for laptops and phones, text descriptions usually give us the specification information of the products such as the model, memory, and storage, etc., which can hardly be obtained from images, while these specification information are the most critical features for price prediction. 

In an overall view, text descriptions play a much more important role in second-hand item price prediction than images. However, the importance of text and image can be quite different for different second-hand items. To check this point, we extract visual feature and textual feature from image and text, and concatenate them together directly without the attention mechanism, then we use the concatenated visual and textual features along with the statistical features to perform price prediction under the percentile constraint (percent constraint on positive items is 60\%) and threshold constraint (threshold value $\epsilon$ in Eq.~\ref{Eq:epsilon} is 0.15), respectively. By comparing the `Ours W/O attention' rows with the `Ours' rows in Table~\ref{tab:baseline}, we can see that combining the visual and textual features with attention mechanism can help improve the performance of price prediction when comparing with feature concatenation without attention mechanism. The reason is that the quality of text descriptions and images are various for different second-hand items. For some items, the uploaded images may be with bad qualities such as blur and uneven illumination, while for some others, the text description may only contain very limited information. Thus, text and image play various important roles in predicting prices for different second-hand item, and by putting attention weights on features, the model can capture the most important information of each second-hand item for price prediction.

When diving deeper into Table~\ref{tab:baseline}, it seems that image feature is not important for price prediction (metrics of `Ours W/O image' are only slightly higher than metrics of `Ours'). However, this is not true. Considering that labeling less positive items will decrease the metric values (imagine an extreme case that the classification model only picks one item with very high-quality image and text as a positive item, then the regression model predict price only for this positive item, the metric values will be very low), we should combine the number of positive items and the metric values together to evaluate the price prediction performance of the proposed system. We need to notice that when performing price prediction without image feature, the binary classification model labels less items as positive (especially in the threshold constraint scenario), and the metrics of prediction with image are slightly lower than the ones of prediction without image. This demonstrates that image feature does benefit the price prediction accuracy, and also the goals of maximizing gain for sellers and promoting transactions for the platform.

\section{Conclusion}

In this paper, we present a multi-modal price suggestion system to provide price suggestions for online second-hand items based on the uploaded item images and text descriptions, which can help the sellers set reasonable listing prices for the items. In the proposed system, a binary classification model is designed to determine whether the uploaded image and text of a second-hand item are qualified for reasonable price suggestion, and a regression model is developed to provide suggested price for items with qualified images and text descriptions. Different constraints are put into the joint training of the classification model and the regression model, so that various demands from the platform operator can be satisfied. Moreover, a customized loss function is designed to optimize the regression model to offer price suggestions which can not only maximize the gain of the seller but also promote the transaction of the second-hand items. Exclusive experiment results on a large real-world dataset demonstrate the effectiveness of the proposed price suggestion system.

{\small

}

\end{document}